\documentclass[10pt,twocolumn,letterpaper]{article}

\usepackage{wacv}
\usepackage{times}
\usepackage{epsfig}
\usepackage{graphicx}
\usepackage{amsmath}
\usepackage{amssymb}
\usepackage{booktabs}
\usepackage{xcolor}         
\usepackage{mathtools}
\usepackage{threeparttable,booktabs}
\usepackage{multirow}
\usepackage{subcaption}
\usepackage{enumitem}

\newcommand{\sref}[1]{Sec.~\ref{#1}}
\newcommand{\eqqref}[1]{Eq.~\ref{#1}}
\newcommand{\figref}[1]{Fig.~\ref{#1}}

\newcommand{\tabref}[1]{Tab~\ref{#1}}

\def\wacvPaperID{1910} 

\wacvalgorithmstrack   

\wacvfinalcopy 

\ifwacvfinal
\usepackage[breaklinks=true,bookmarks=false]{hyperref}
\else
\usepackage[pagebackref=true,breaklinks=true,colorlinks,bookmarks=false]{hyperref}
\usepackage[accsupp]{axessibility}  

\fi

\begin{document}

\title{Learning Classifiers of Prototypes and Reciprocal Points \\
for Universal Domain Adaptation}

\author{Sungsu Hur\qquad Inkyu Shin\qquad Kwanyong Park\qquad Sanghyun Woo\qquad In So Kweon\\
KAIST\\
}

\maketitle
\thispagestyle{empty}

\begin{abstract}

Universal Domain Adaptation aims to transfer the knowledge between the datasets by handling two shifts: domain-shift and category-shift. The main challenge is correctly distinguishing the unknown target samples while adapting the distribution of known class knowledge from source to target. Most existing methods approach this problem by first training the target adapted known classifier and then relying on the single threshold to distinguish unknown target samples. However, this simple threshold-based approach prevents the model from considering the underlying complexities existing between the known and unknown samples in the high-dimensional feature space. In this paper, we propose a new approach in which we use two sets of feature points, namely dual \textbf{C}lassifiers for \textbf{P}rototypes and \textbf{R}eciprocals (\textbf{CPR}). Our key idea is to associate each prototype with corresponding known class features while pushing the reciprocals apart from these prototypes to locate them in the potential unknown feature space. The target samples are then classified as unknown if they fall near any reciprocals at test time. 
To successfully train our framework, we collect the partial, confident target samples that are classified as known or unknown through on our proposed multi-criteria selection. We then additionally apply the entropy loss regularization to them. For further adaptation, we also apply standard consistency regularization that matches the predictions of two different views of the input to make more compact target feature space. 
We evaluate our proposal, CPR, on three standard benchmarks and achieve comparable or new state-of-the-art results.
We also provide extensive ablation experiments to verify our main design choices in our framework.
\end{abstract}

\section{Introduction}

Deep-learning based approaches have shown remarkable success on recognition tasks~\cite{he2016deep,krizhevsky2012imagenet,simonyan2014very} given a huge amount of data, but do not generalize well to the data from newly seen domain. Therefore, labeled datasets for the novel domain need to be constructed, which requires tremendous labeling efforts in time and cost. Unsupervised Domain Adaptation (UDA) addresses this problem by 
handling the domain shift from labeled source data to unlabeled target data. However, conventional UDA methods~\cite{tzeng2017adversarial,hoffman2018cycada,long2018conditional,saito2018maximum} only perform when the both domains share the label space, which limits applicability when the category shift happens. In that sense, several DA scenarios have recently proposed a more practical perspective that takes into account both domain shift and category shift during the domain adaptation: Open-set Domain Adaptation (OSDA)~\cite{panareda2017open,saito2018open} and Partial Domain Adaptation (PDA)~\cite{cao2018partial}. OSDA assumes there are target private classes that are not shown in source domain. PDA deals with a vice versa scenario where source domain possesses its own classes. However, their settings are out of line with the real-world difficulty where we cannot know how the label space between two domains is different in advance.
To make up for this, Universal DA (UniDA)~\cite{you2019universal} has been introduced to account for the uncertainty about the category-shift between source and target domains. 
The purpose of the UniDA is to make a model that is applicable to any category shift scenarios and classifies the target samples into either one of the correct known classes or the unknown classes. The main challenge for UniDA is to detect unknown samples correctly while transferring domain knowledge from source domain to the target domain. 

Early works attempted to solve the issues with following techniques: calculating unknown scores with domain similarity and entropy value~\cite{you2019universal}, employing multiple uncertainties to decide unknown samples~\cite{fu2020learning}, proposing a neighborhood clustering techniques with entropy optimization for rejecting unknown categories~\cite{saito2020universal}. All of these methods manually set a threshold to determine the label space of target samples. OVANet~\cite{saito2021ovanet} deals with this limitation by adopting an additional One-vs-All classifier that aims to find an adaptive threshold between known and unknown classes. 
Despite of their efforts, they still lack the ability to capture the distinctive properties of known and unknown samples. Moreover, they heavily rely on single criteria (threshold) for dividing the target samples into known and unknown, which is not powerful enough to handle the category shift in real-world. Those two limitations eventually lead to downgrade the performance of not only detecting unknown samples but also adapting known classes between source and target domains.

Motivated by the above limitations in previous methods for UniDA, we propose to explicitly learn feature characteristics of both known and unknown samples with newly proposed dual \textbf{C}lassifiers for \textbf{P}rototypes and \textbf{R}eciprocals (\textit{\textbf{CPR}}).
Along with standard prototype learning~\cite{snell2017prototypical} to represent known classes, we adopt the concept of reciprocals~\cite{chen2021adversarial} to symbolize unknown samples.
Considering the complexities from mingled domain and category shift, the reciprocal points discover the unknown feature spaces in curriculum manner.
At the warm-up phase, the reciprocal points are first initialized at unexploited regions from known source classes, where the unknown classes potentially place in.
At the same time, the domain shift is gradually reduced by consistency regularization.
The target samples are augmented in the weak and strong views and consistency between the predictions of the two views is increased.

After warm-up, the dual classifiers have better representation power to distinguish between known and unknown samples regardless of domain.
To faithfully utilize it, we collect confident known/unknown samples to regularize the both classifiers.
To this end, we propose carefully designed multiple criteria to filter the samples, considering the natural properties of dual classifiers.
Given the filtered known/unknown samples, corresponding prototypes/reciprocals are close to them, respectively.
By doing so, the reciprocals explicitly locate the unknown target classes.
With our novel dual classifiers and training recipes, the feature distribution of source/target samples are aligned and each classifiers successfully identify both known and unknown samples.

Here are our main contributions:
\begin{enumerate}
\vspace{-1mm}
\setlength\itemsep{0.3em}
    \item We propose \textbf{\textit{CPR}}, a universal domain adaptation framework with dual classifiers including learnable unknown detector called reciprocal classifier. With the help of newly proposed objective function, it can achieve to capture both known and unknown feature space. 
    
    \item We devise a new multiple criteria to find more reliable samples for both known and unknown classes, considering the natural structure of feature space extracted from dual classifiers and their confidence thresholds.
    
    \item We demonstrate our novel framework under different universal domain adaptation benchmarks with extensive ablation studies and experimental comparisons against the previous state-of-the art methods. 
\end{enumerate}

\begin{figure*}[t!]
    \centering
    \includegraphics[width=0.9\linewidth]{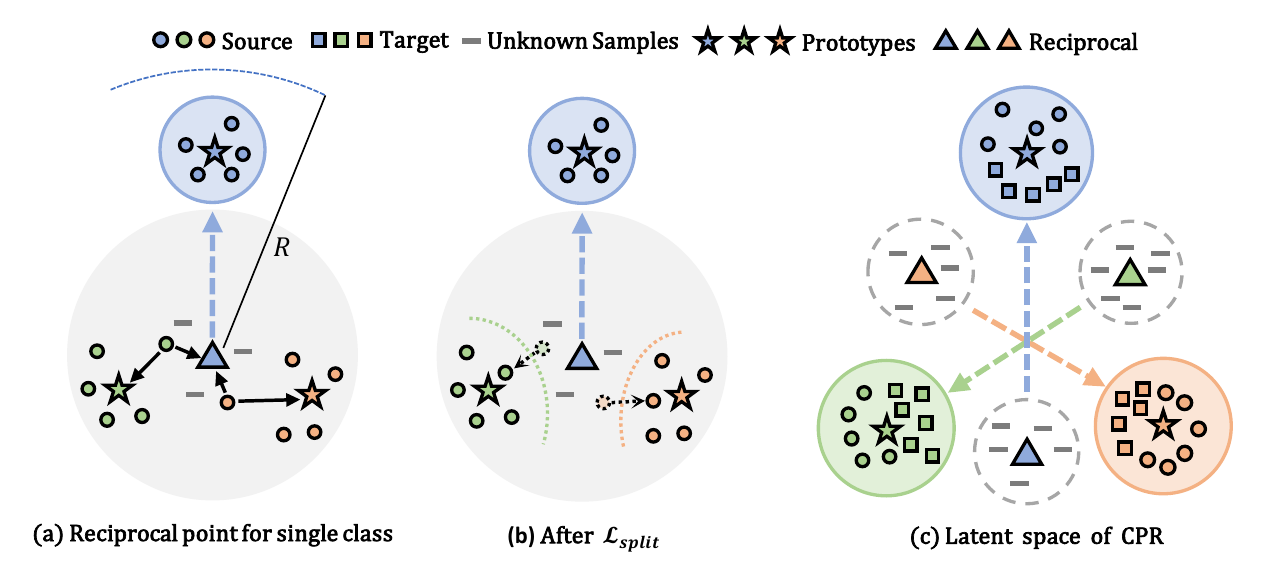}
    \caption{Dual classifiers are initially trained with labeled source samples. To ensure reciprocal points and prototypes are distinct enough, we devise split loss and further minimize weighted entropy loss to make target samples confident regardless of whether those are known or unknown samples.}
    \label{fig1}
\end{figure*}

\section{Related work}

\textbf{Unsupervised Domain Adaptation.} The main purpose of Unsupervised Domain Adaptation(UDA) is to transfer the knowledge from source to target domain while accounting for domain shift between them. A closed-set DA(CDA) is the conventional UDA setting where two domains share the same label space.  Methods utilizing adversarial learning~\cite{saito2018maximum, zhang2019domain, li2019cycle} or self-training~\cite{liang2021domain, zou2018unsupervised} with generated pseudo labels of target samples have been proposed to solve closed-set DA. However, this scenario does not perform when category-shift happens between the datasets. 
Motivated by this limitation, several scenarios in UDA have been proposed to handle category-shift.
Among them, Partial DA(PDA) deals with the case with presence of private source classes. To solve this task, Most methods design weighting schemes to re-weight source examples during domain alignment~\cite{cao2018partial, liang2020balanced, zhang2018importance}. Open-set DA(OSDA) is another approach to handle private target classes that is never seen on source domain~\cite{panareda2017open, kundu2020towards, saito2018open}.

\textbf{Universal Domain Adaptation.} 
All of aforementioned methods only focus on their fixed category shift scenario, but in reality we mostly could not access to prior knowledge of label space relationship between source and target domain. Universal Domain Adaptation (UniDA) have been proposed to address the issue. UAN~\cite{you2019universal} first introduced UniDA framework, which utilizes a weighting mechanism to discover label sets shared by both domains. CMU~\cite{fu2020learning} further improved measure of uncertainty to find target unknown classes more accurately. DANCE~\cite{saito2020universal} learns the target domain structure by neighborhood clustering, and used an entropy separation loss to achieve feature alignment. Recently, OVANet~\cite{saito2021ovanet} designed one-vs-all classifier to obtain unknown score and adopt an adaptive threshold. However, their single threshold methods fail to explicitly bring out the unknown features from the target samples.
To address the above weakness, we adopt a novel dual-classifier framework for prototype and reciprocal to detect the properties of known and unknown samples separately with multi-criteria selection.

\textbf{Open set recognition.} 
~\cite{scheirer2012toward} defined Open set recognition(OSR) problem for the first time and proposed a base framework to perform training and evaluation. With rapid development of deep neural networks, ~\cite{bendale2016towards} incorporated deep neural networks into OSR by introducing the OpenMax function. Then both ~\cite{ge2017generative} and ~\cite{neal2018open} tried to synthesize training samples of unseen classes via the Generative Adversarial Network. Since ~\cite{yang2020convolutional} attempted to combine prototype learning with deep neural networks for OSR, they achieved the new state-of-the art. Prototypes refer to representative samples or latent features for each class. ~\cite{yang2020convolutional} introduced Convolutional Prototype Network (CPN), in which prototypes per class were jointly learned during training. ~\cite{chen2021adversarial, chen2020learning} learned discriminative reciprocal points for OSR, which can be regarded as the inverse concept of prototypes. In this paper, we incorporate a reciprocal points as learnable representation points to differentiate ``known" and potential ``unknown" samples in UniDA. 

\section{Methodology}
In UniDA, there exist a labeled source domain $\mathcal{D}_s=\{(x_i^s, y_i^s)\}_{i=1}^{N_s}$ with closed (known) categories $L_s$ and an unlabeled target domain $\mathcal{D}_t=\{(x_i^t)\}_{i=1}^{N_t}$ with categories $L_t$ that could be partially overlapped with $L_s$ and potentially consists of open (unknown) classes.
$L_s$ and $L_t$ denote the label sets of the source domain and target domain, respectively. Our goal is to label the target samples with either one of the known labels $C_s$ or the ``unknown" label.

\begin{figure*}[t!]
    \centering
    \includegraphics[width=0.8\linewidth]{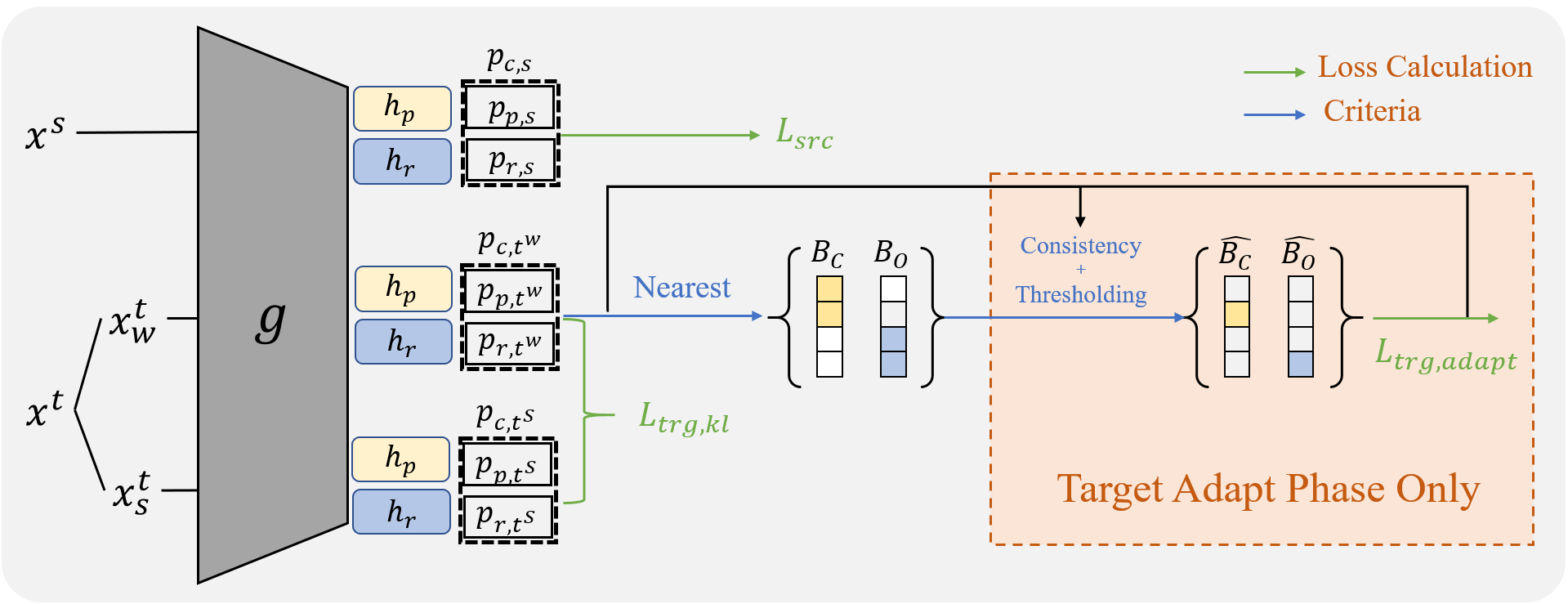}
    \caption{Overview of Network. During the whole training time, we use source domain to associate each prototype with corresponding known class features while pushing the reciprocals apart from these prototypes $\mathcal{L}_{src}$ (\sref{sec3.1}, \sref{sec3.2}).
    On the contrary, we adopt curriculum learning for target domain as follows:
    We first guide the model to gradually adapt from source to target via standard consistency regularization $\mathcal{L}_{trg,kl}$ (\textbf{warm-up phase}).
    Then, in the \textbf{adaptation phase}, we apply additionally enforce regularization $\mathcal{L}_{trg,adapt}$ on the target samples that are classified as known or unknown based on the proposed multiple criteria.
    }
    \label{fig2}
\end{figure*}

\paragraph{Overview.} 
As prototypes are the points describing the characteristics of corresponding known class, other reciprocal points are required to help model interpret unknown feature correctly. In that sense, reciprocal points~\cite{chen2020learning} are utilized to symbolize the unknown feature space while also prototypes are used to point out known feature space at the same time, which motivates us to develop a dual-classifier framework for them. Furthermore, we introduce multi-criteria selection mechanism to effectively adapt the model to the target distribution with the confident target samples.
As shown in \figref{fig2}, our model consists of a shared feature extractor $g$ and two classifiers, prototype classifier $h_p$ and reciprocal classifier $h_r$. Feature extractor $g$ takes an input $x$ and outputs a feature vector $f=g(x)$. Two classifiers $h_p$ and $h_r$ consist of weight vectors $\left[\boldsymbol{p_1}, \boldsymbol{p_2}, \dots, \boldsymbol{p_K}\right]$ and $\left[\boldsymbol{r_1}, \boldsymbol{r_2}, \dots, \boldsymbol{r_K}\right]$ that indicates corresponding unnormalized output logits of $K$ known classes. 
We conduct the softmax function to obtain prototypical probability $p_{p}=Softmax(h_p(f)) \in \mathbb{R}^{K}$ and reciprocal probability $p_{r}=Softmax(h_r(f)) \in \mathbb{R}^{K}$. We also define collaborative probability $p_{c} = Softmax([h_p(f),h_r(f)]) \in \mathbb{R}^{2K}$ where $[,]$ means the concatenation of two logits. 
For the source domain, while each classifier is trained to predict correct ground-truth label of input like using two different classifiers~\cite{saito2018maximum}, we ensure reciprocal points become effectively far away from known source data by using a new margin loss (\sref{sec3.1} and \sref{sec3.2}). For the target samples, we firstly augment them with two different views ($x^{t}_s$ and $x^{t}_w$ as strong and weak views respectively). And then, we make the model to be aware of target distribution by giving consistency regularization on collaborative probability $p_{c}$ for two views in the first phase (\sref{sec3.3.1}). In the next phase, we firstly separate total input batch into $B_C$ and $B_O$ by pseudo labels.
\begin{equation}
    x^t \in \begin{dcases}
        B_c & argmax(p_c) < K \\
        B_o & argmax(p_c) \geq K
    \end{dcases}
\end{equation}
They are further filtered out to detect confident known and unknown samples based on the multi-criteria selection mechanism (\sref{sec3.3.2}). Last but not least, using trained dual classifiers, we classify the target samples into proper known classes while filtering out unknown samples with $p_c$ in inference (\sref{sec3.4}). 

\subsection{Preliminary: Reciprocal points for classification}\label{sec3.1}
Before discussing the proposed framework, we first review reciprocal points introduced in ~\cite{chen2020learning, chen2021adversarial}. The main idea is to learn latent representation points to be the farthest ones from the corresponding classes, which is the reverse concept of prototypes. Given source feature vectors $f_s$, the classifier for reciprocal is trained by minimizing the reciprocal points classification losses based on the negative log-probability of the true class k:
\begin{gather}
    d(f_s, \boldsymbol{r_k}) = -f_s \cdot\boldsymbol{r_k} \\
    p_{r}\left(y=k\vert f_s,h_{r}\right) = \frac{e^{d(f_s, \boldsymbol{r_k})}}{\sum^{K}_{i=1}e^{d(f_s,\boldsymbol{r_i})}} \label{eq1} \\
    \mathcal{L}_{CE_{r}} = -\log p_{r}\left(y=k\vert f_s,h_r\right) \label{eq2}
\end{gather}
where $d$ is a distance metric. In this paper, we simply apply minus dot product to estimating distance. 
Although ~\eqqref{eq1} helps $h_{r}$ to maximize the distance between reciprocal points and corresponding samples, extra class space including infinite unexploited unknown space can be expanded with no restriction. To separate unknown space with known one as much as possible, the open space should be restricted.
To restrict unknown space, ~\cite{chen2020learning, chen2021adversarial} additionally propose to minimize the following loss given a feature vector $f_s$ from category $k$
\begin{equation} \label{eq3}
    \mathcal{L}_{o} = max(d(f_s, \boldsymbol{r_k})-R,0)
\end{equation}
where $R$ is a learnable margin. As shown in ~\figref{fig1}a, by limiting the distance $d(f_s, \boldsymbol{r_k})$ less than $R$, the distance to the remaining samples of extra classes would be also reduced indirectly less than $R$. In other words, the open space risk can be implicitly bounded by utilizing ~\eqqref{eq3} and features from unknown classes could be congregated around reciprocal points. In light of this, we interpret reciprocal points as potential representation points for target private classes. 

\subsection{Learning from Source Domain}\label{sec3.2}
As explained, the reciprocal classifier can be trained by minimizing ~\eqqref{eq2}. Given the dual classifiers, the prototype classifier can be learned by minimizing the following standard cross-entropy loss, 
\begin{gather}
    p_{p}\left(y=k\vert f_s,h_p\right) = \frac{e^{-d(f_s, \boldsymbol{p_k})}}{\sum^{K}_{i=1}e^{-d(f_s,\boldsymbol{p_k})}} \label{eq4} \\
    \mathcal{L}_{CE_p} = -\log p_{p}\left(y=k\vert f_s,h_p\right) \label{eq5}
\end{gather}
This loss would make prototypes to be close to the correct samples. Moreover, ~\eqqref{eq3} should be taken to restrict unknown space and make features be more compact. However, if we naively minimize these losses, we cannot handle the case where some reciprocal points are getting closer to features of other known classes as shown in the ~\figref{fig1}a. 
To fix this, each source feature should be more closer to its prototype than any other reciprocal points. 
By choosing a nearest reciprocal point as reference, we can effectively and clearly separate known space from unknown one as shown in \figref{fig1}b. Hence, the proposed objective for the true class k is denoted as follows:
\begin{equation}
    \mathcal{L}_{split} = max(d(f_s, \boldsymbol{p_k})-\mathop{min}_{i}(d(f_s, \boldsymbol{r_i})),0)
\end{equation}

Then, the overall training loss for source domain can be computed as follows:
\begin{equation}\label{eq7}
    \mathcal{L}_{src} = \mathcal{L}_{CE_p}+\mathcal{L}_{CE_r}+\lambda (\mathcal{L}_{o}+\mathcal{L}_{split})
\end{equation}
This loss allows both classifiers to classify known class, while at the same time allowing the reciprocal point to be distributed in unknown feature space.
\subsection{Learning from Target Domain}\label{sec3.3}
As described in the overview, unknown samples are found based on whether their closest point is the one of reciprocal points or not. 
However, if we proceed with the learning before reciprocal points are placed in unknown feature space, there would be a huge performance degradation due to immature function to differentiate known and unknown class. Even though reciprocal points are clearly separated from prototypes, noisy samples
are inevitable due to the domain gap between source and target domains. Thus, It is essential to give reciprocal points enough time to be separated from known class space and effectively detect reliable unknown feature.
To solve these issues, the training is done in curriculum manner from warm-up phase($iters<i_{w}$) to adaptation phase($iters\geq i_{w}$). In the warm-up phase, reciprocal points are gradually aligned to the open space and thresholds for selecting confident samples are updated in online manner. Then, in the adaptation phase, we minimize the weighted entropy loss with samples selected through well calibrated multi criteria including thresholds and the consistency between dual classifiers.

\subsubsection{Warm-up Phase}\label{sec3.3.1}
As shown in ~\figref{fig2} there are two different views for the target domain, weak augmented view $x_w^t$ and strong augmented view $x_s^t$. The model is trained with the $p_c$ of two views to become similar to generate more compact target features. 
\begin{equation}\label{eq8}
    \mathcal{L}_{kl} = KL\left(p_{c}(x_s^t)\Vert p_{c}(x_w^t)\right)
\end{equation}

During the warm-up phase, the model is trained using Eq~\ref{eq7} and Eq~\ref{eq8} to generate compact target features with well initialized points. Along with this, thresholds for known and unknown classes should be calculated to select reliable samples in the adaptation phase. Two thresholds are progressively updated by moving average of mean collaborative probability  of $B_c$ and $B_o$
\begin{align}
    \rho_c&=\alpha*\rho_c+(1-\alpha)*\mathop{\mathbb{E}}_{x^t\in B_c}max(p_{c}) \\
    \rho_o&=\alpha*\rho_o+(1-\alpha)*\mathop{\mathbb{E}}_{x^t\in B_o}max(p_{c})
\end{align}
where $\rho_c$ and $\rho_o$ are initially set as 0. These thresholds would be used in the adaptation phase to select more confident samples and continue to be updated according to the model adapting to the target domain. 

\subsubsection{Adaptation Phase}\label{sec3.3.2}
In the adaptation phase, given the warm-up model and thresholds, we additionally enforce entropy regularization on the target samples that pass multi-criteria we propose. We detail the criteria below.

\paragraph{Multi-Criteria for Selection.}
By design of the framework, index of the nearest prototype and that of the farthest reciprocal point should be same for known classes. 
This naturally motivates us to design the first criteria of examining whether the same index of the classifiers of prototypes and reciprocals are fired in distinguishing the known and unknown classes.
Also, we evaluate if the $max(p_c)$ is above the threshold to obtain only the confident predictions. 
We note that the threshold values, $\rho_c$ and $\rho_o$, are continuously updated in the adaptation stage.
By putting together, confident and reliable sets $\hat{B_c}$ and $\hat{B_o}$ are sampled from $B_c$ and $B_o$ respectively as shown in \figref{fig2}.
\begin{align*}
    \hat{B_c}:  max(p_{c})\geq \rho_c\quad \&\quad argmax(p_{p})=argmax(p_{r})\\
    \hat{B_o}:  max(p_{c})\geq \rho_o\quad \&\quad argmax(p_{p})\neq argmax(p_{r})
\end{align*}
Since the output of weak augmented view is more reliable than strong augmented one, we first get the selected weak augmented samples. Then, we also take account for the strong augmented samples which are pairs of selected weak augmented ones. After that, for the strong augmented view, we forward it one more condition, judging whether its nearest point is the same with that of weak augmented view.
By following the above multi-criteria, we could send more confidently encoded features to the dual classifiers. 

\paragraph{Weighted Entropy.} From the selected samples, we could calculate entropy $H(x^t)=-p_c(x^t) \log p_c(x^t)$. By minimizing the entropy, selected samples become more closer to their nearby points and more confident. However, this vanilla entropy minimization may leads model being biased to either known or unknown classes due to class imbalance. Hence, we weight each entropy of known and unknown classes using the number of selected samples as follows. 
\begin{equation}
    \mathcal{L}_{ent}=w*\mathop{\mathbb{E}}_{x^t\in \hat{B_c}}H(x^t) + (1-w)*\mathop{\mathbb{E}}_{x^t\in  \hat{B_o}}H(x^t)
\end{equation}
where $ w=\frac{\left\vert \hat{B_o}\right\vert}{\left\vert \hat{B_c}\right\vert+\left\vert  \hat{B_o}\right\vert}$. Furthermore, we also add Eq~\ref{eq3} to minimize open feature space using detected pseudo known target samples $\hat{B_c}$. Consequently, the overall training loss can be computed as follows:

\begin{gather}
    \mathcal{L}_{trg}=\begin{dcases}
        \mathcal{L}_{kl} & iters < i_{w} \\
        \mathcal{L}_{kl} + \mathcal{L}_{ent} +\lambda\mathcal{L}_o & iters \geq i_{w} 
    \end{dcases} \\
    \mathcal{L}_{all} = \mathcal{L}_{src}+\mathcal{L}_{trg}
\end{gather}

\begin{table*}[t]
\caption{H-score of each method on \textbf{Office-Home} for OSDA.}   
\label{osda2}
\centering
{\scriptsize
\begin{tabular}{l |c c c c c c c c c c c c | c}
\toprule[1.0pt]
\multirow{2}{*}{Method} & \multicolumn{12}{c|}{OfficeHome (25/0/40)} & \\
 & Ar$\to$ Cl & Ar$\to$ Pr & Ar$\to$ Re & Cl$\to$ Ar & Cl$\to$ Pr & Cl$\to$ Re & Pr$\to$ Ar & Pr$\to$ Cl & Pr$\to$ Re & Re$\to$ Ar & Re$\to$ Cl & Re$\to$ Pr & Avg \\
\hline
ROS & 60.1 & 69.3 & 76.5 & 58.9 & 65.2 & 68.6 & 60.6 & 56.3 & 74.4 & 68.8 & 60.4 & 75.7 & 66.2  \\
UAN  & 0.0 & 0.0 & 0.2 & 0.0 & 0.2 & 0.2 & 0.0 & 0.0 & 0.2 & 0.2 & 0.0 & 0.1 & 0.1  \\
CMU  & - & - & - & - & - & - & - & - & - & - & - & - & -  \\
DCC  & 56.1 & 67.5 & 66.7 & 49.6 & 66.5 & 64.0 & 55.8 & 53.0 & 70.5 & 61.6 & 57.2 & 71.9 & 61.7 \\
OVA  & 58.4 & 66.3 & 69.3 & 60.3 & 65.1 & 67.2 & 58.8 & 52.4 & 68.7 & 67.6 & 58.6 & 66.6 & 63.3 \\
\hline
CPR & 57.1 & 67.2 & 75.7 & 64.9 & 66.8 & 65.6 & 64.5 & 57.3 & 73.8 & 71.0 & 60.9 & 74.4 & \textbf{66.6} \\ 
\hline
\bottomrule
\end{tabular}}

\end{table*}
\begin{table}[h!]
\caption{H-score of each method on \textbf{Office} and \textbf{VisDA} for OSDA.}   
\label{osda1}
\centering
\resizebox{\linewidth}{!}{
\begin{tabular}{l|c c c c c c | c | c}
\toprule[1.0pt]
\multirow{2}{*}{Method}& \multicolumn{6}{c|}{Office (10/0/21)} &  & VisDA \\
      & A$\to$ D & A$\to$ W & D$\to$ A & D$\to$ W & W$\to$ A & W$\to$ D & Avg & (6/0/6) \\
\hline
ROS & 65.8 & 71.7 & 87.2 & 94.8 & 82.0 & 98.2 & 83.3 & 50.1  \\
UAN  & 38.9 & 46.8 & 68.0 & 68.8 & 54.9 & 53.0 & 55.1 & 51.9  \\
CMU  & - & - & - & - & - & - & - & -  \\
DCC  & 58.3 & 54.8 & 67.2 & 89.4 & 85.3 & 80.9 & 72.6 & 70.7 \\
OVA  & 90.5 & 88.3 & 86.7 & 98.2 & 88.3 & 98.4 & \textbf{91.7} & 53.5 \\
\hline
CPR & 90.4 & 89.4 & 86.7 & 98.5 & 88.6 & 92.7 & 91.1 & \textbf{79.4} \\ 
\hline
\bottomrule
\end{tabular}}

\end{table}

\subsection{Inference}\label{sec3.4}
In the test phase, we simply use the collaborative probability $p_{c}$ to see what is the nearest point. If the nearest point is one of prototypes, it is classified as corresponding known class, and if it is one of reciprocal points, it is classified as unknown class.

\section{Experiments}
\subsection{Setup}
\paragraph{Datasets.} We conduct experiments on three datasets. \textbf{Office-31}~\cite{saenko2010adapting} consists of 4652 images in 31 categories from three distinct domains: DSLR (D), Amazon (A), and Webcam (W). The second benchmark dataset \textbf{OfficeHome}~\cite{venkateswara2017deep} is a more challenging one, which contains 15500 images with 65 classes and four domains Art (Ar), Clipart(Cl), Product(Pr), and Real-World (Re). The third dataset \textbf{VisDA}~\cite{peng2017visda} is a large-scale dataset, where the source domain contains 15K synthetic images and the target domain consists of 5K images from the real world. Let $\left\vert L_s\cap L_t\right\vert$, $\left\vert L_s- L_t\right\vert$ and $\left\vert L_t- L_s\right\vert$ denote the number of common categories, source private categories and target private categories, respectively. Following ~\cite{saito2020universal}, we split the classes of each benchmark and show the split of each experimental setting in a corresponding table. 

\paragraph{Evaluation Metric.}
Following ~\cite{saito2021ovanet}, we evaluate the performance using H-score for both OSDA and UniDA. H-score is the harmonic mean of the accuracy on common classes($acc_c$) and accuracy on the ``unknown" classes($acc_t$) as:
\begin{equation}
    H_{score}=\frac{2acc_c \cdot acc_t}{acc_c+acc_t}
\end{equation}
The H-score is high only when both the ``known" and ``unknown" accuracies are high. Thus, H-score accurately measures both accuracies.

\begin{table*}[t!]
\caption{H-score of each method on \textbf{Office-Home} for UniDA.}   
\label{opda2}
\centering
{\scriptsize
\begin{tabular}{l |c c c c c c c c c c c c | c}
\toprule[1.0pt]
\multirow{2}{*}{Method}  & \multicolumn{12}{c|}{OfficeHome (10/5/50)} & \\
 & Ar$\to$ Cl & Ar$\to$ Pr & Ar$\to$ Re & Cl$\to$ Ar & Cl$\to$ Pr & Cl$\to$ Re & Pr$\to$ Ar & Pr$\to$ Cl & Pr$\to$ Re & Re$\to$ Ar & Re$\to$ Cl & Re$\to$ Pr & Avg \\
\hline
UAN   & 51.6 & 51.7 & 54.3 & 61.7 & 57.6 & 61.9 & 50.4 & 47.6 & 61.5 & 62.9 & 52.6 & 65.2 & 56.6  \\
CMU   & 56.0 & 56.9 & 59.2 & 67.0 & 64.3 & 67.8 & 54.7 & 51.1 & 66.4 & 68.2 & 57.9 & 69.7 & 61.6  \\
DCC   & 58.0 & 54.1 & 58.0 & 74.6 & 70.6 & 77.5 & 64.3 & 73.6 & 74.9 & 81.0 & 75.1 & 80.4 & 70.2 \\
OVA   & 62.8 & 75.6 & 78.6 & 70.7 & 68.8 & 75.0 & 71.3 & 58.6 & 80.5 & 76.1 & 64.1 & 78.9 & 71.8 \\
\hline
CPR  & 59.0 & 77.1 & 83.7 & 69.7 & 68.1 & 75.4 & 74.6 & 56.1 & 78.9 & 80.5 & 63.0 & 81.0 & \textbf{72.3} \\ 
\hline
\bottomrule
\end{tabular}}
\end{table*}

\begin{table}[t!]
\caption{H-score of each method on \textbf{Office} and \textbf{VisDA} for UniDA.}   
\label{opda1}
\centering
\resizebox{\linewidth}{!}{
\begin{tabular}{l |c c c c c c | c|c}
\toprule[1.0pt]
\multirow{2}{*}{Method}& \multicolumn{6}{c|}{Office (10/10/11)} &  & VisDA \\
      & A$\to$ D & A$\to$ W & D$\to$ A & D$\to$ W & W$\to$ A & W$\to$ D & Avg  & (6/3/3) \\
\hline
ROS  & 71.4 & 71.3 & 81.0 & 94.6 & 95.3 & 79.2 & 82.1 & 50.1  \\
UAN   & 59.7 & 58.6 & 60.1 & 70.6 & 60.3 & 71.4 & 63.5 & 30.5  \\
CMU   & 68.1 & 67.3 & 71.4 & 79.3 & 80.4 & 72.2 & 73.1 & 34.6  \\
DCC   & 88.5 & 78.5 & 70.2 & 79.3 & 88.6 & 75.9 & 80.2 & 43.0 \\
OVA   & 85.8 & 79.4 & 80.1 & 95.4 & 94.3 & 84.0 & 86.5 & 53.1 \\
\hline
CPR  & 84.4 & 81.4 & 85.5 & 93.4 & 91.3 & 96.8 & \textbf{88.8} & \textbf{58.2} \\ 
\hline
\bottomrule
\end{tabular}}

\end{table}
\paragraph{Implementation Details.}
We use ResNet50~\cite{he2016deep} pre-trained on ImageNet~\cite{deng2009imagenet} as our backbone network following previous works. The batch size is set to 36, and the hyperparameters $\lambda$ and $\alpha$ are set as 0.1 and 0.99, respectively. We adopt horizontal flip and random crop as weak augmentation and augmentation used in FixMatch~\cite{sohn2020fixmatch} as strong augmentation. The number of iterations for warm-up phase, $i_{w}$ is set as 1000, where thresholds for all the experiments are saturated. In case of large-scale dataset VisDA, $i_{w}$ is set as $\max (\left\vert\mathcal{D}_s\right\vert, \left\vert\mathcal{D}_t\right\vert)$/(batch size) to allow model to see all the samples in both datasets. Following previous works, We train our model for total 10000 iterations including $i_{w}$. We conduct all experiments with single GTX 1080ti GPU.
\subsection{Main Results}
In this section, we show quantitative evaluations on the aforementioned four benchmark settings by reporting H-score value. For each benchmark setting, we mainly compare our method with the state-of-the-art baselines: ROS~\cite{bucci2020effectiveness}, UAN~\cite{you2019universal}, CMU~\cite{fu2020learning}, DCC ~\cite{li2021domain}, OVANet~\cite{saito2021ovanet}.

\paragraph{Experimental Results}
As seen in \tabref{osda2} and \tabref{osda1}, CPR outperforms or comparable to baseline methods on the OSDA setting. Our method achieves the best H-score 79.4$\%$ and 66.6$\%$ on the large-scale dataset VisDA and OfficeHome, which outperforms the other methods, and second best H-score of 91.0$\%$ on the Office-31. For the large-scale VisDA dataset, CPR gives more than 8$\%$ improvements compared to the other methods. In case of UniDA, CPR also shows superior performance compared to the other methods as shown in \tabref{opda2} and \tabref{opda1}.
In summary, CPR outperforms or comparable to previous state-of-the-art methods across different DA settings. 
It proves that our dual-classifier framework is robustly powerful in several benchmarks with various UniDA and OSDA settings. 

\begin{figure*}
    \subfloat[Art to Product]{{\includegraphics[width=0.24\textwidth ]{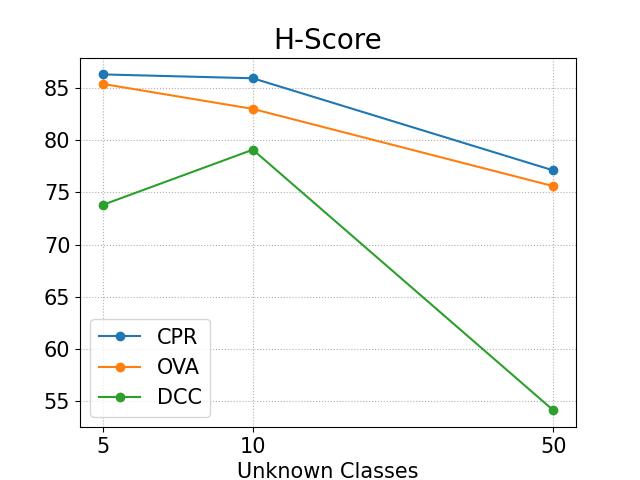} }\label{fig3.1}}
    \subfloat[Real to Product]{{\includegraphics[width=0.24\textwidth ]{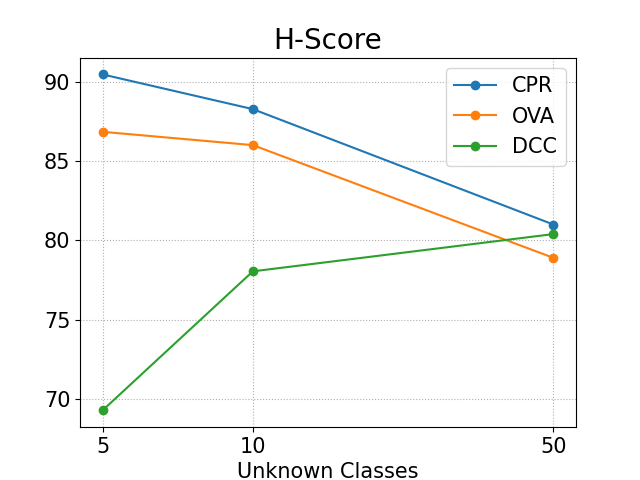} }\label{fig3.2}}
    \subfloat[Product to Art]{{\includegraphics[width=0.24\textwidth ]{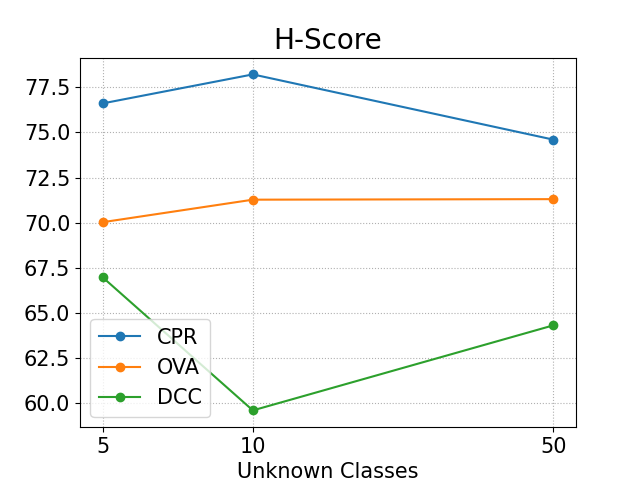} }\label{fig3.3}}
    \subfloat[Vidsa OSDA]{{\includegraphics[width=0.24\textwidth ]{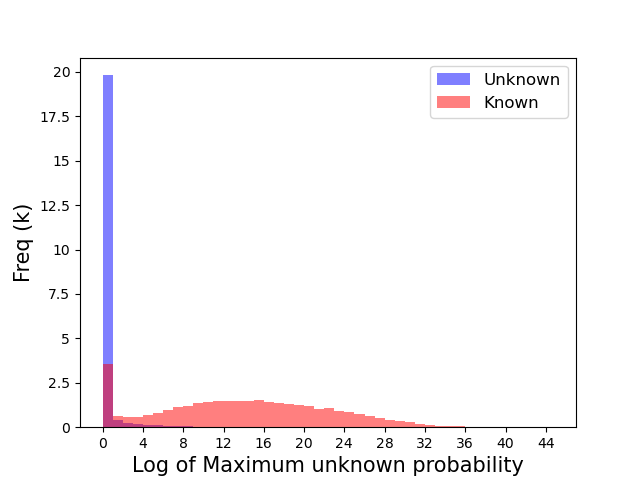} }\label{fig3.4}}
    \caption{(a)$\sim$(c): H-score as varying the number of unknown classes in OfficeHome ($\left\vert L_s\cap L_t\right\vert=10,\, \left\vert L_s - L_t\right\vert=5$) (d): Histogram of log of maximum unknown probability in VisDA}
    \label{fig3}
\end{figure*}

We analyze the behavior of CPR across different number of unknown classes. We perform 4 UniDA experimens in OfficeHome with fixing the number of common classes and source private ones and compare H-score of those with DCC and OVA. \figref{fig3} shows that the performance of CPR is always better than other baseline methods and robustness to the number of unknown classes.

\begin{figure*}[t]
    \subfloat[full CPR]{{\includegraphics[width=0.23\textwidth ]{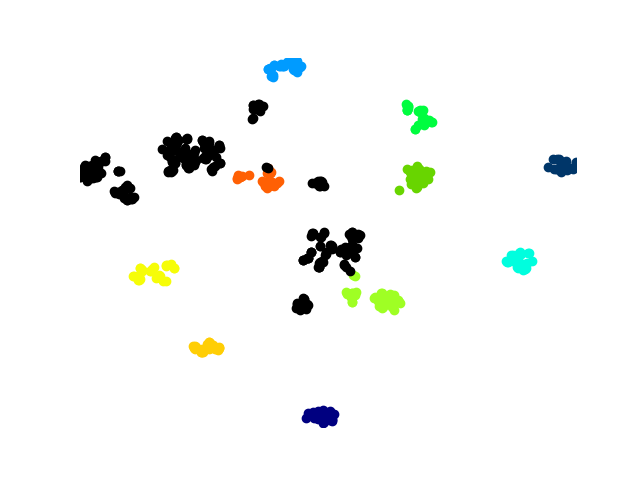} }\label{fig4.1}}%
    \subfloat[w/o $\mathcal{L}_{split}$]{{\includegraphics[width=0.23\textwidth ]{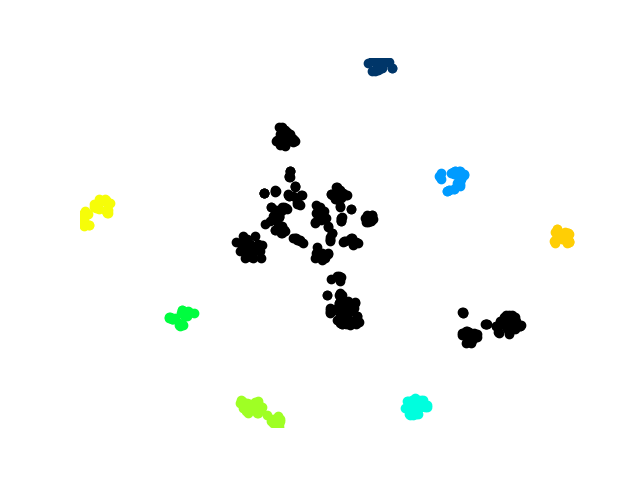} }\label{fig4.2}}%
    \subfloat[w/o consist]{{\includegraphics[width=0.23\textwidth ]{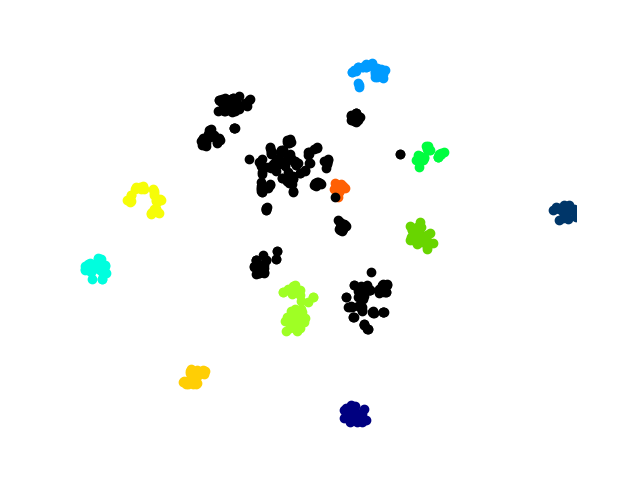} }\label{fig4.3}}%
    \subfloat[w/o thr]{{\includegraphics[width=0.23\textwidth ]{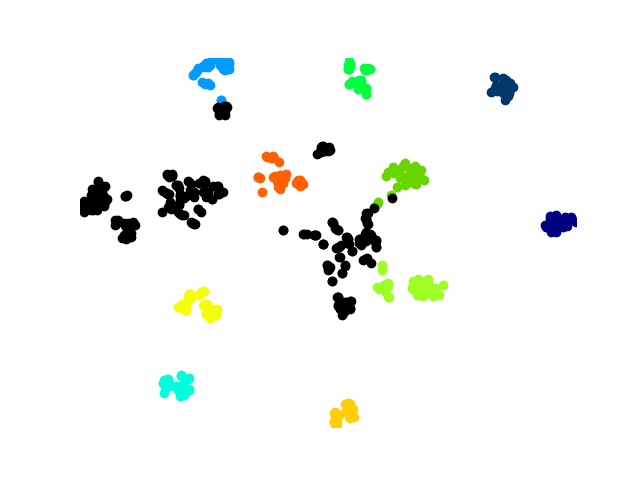} }\label{fig4.4}}%
    
    \caption{Feature visualization on D2W in Office OSDA. Black plots are unknown samples, others are known samples}
    \label{fig4}
\end{figure*}

\begin{table}[t]
    \caption{Results of ablation studies}
    \begin{subtable}[t]{0.5\textwidth}
        \centering
        \caption{Analysis on $\mathcal{L}_{split}$ and Warm-up stage }
        {\footnotesize
        \begin{tabular}{c|c c}
            \toprule
            Ablation Study & Office(OSDA) & VisDA(OSDA)\\
            \hline
            w/o $\mathcal{L}_{split}$ & 59.55 & 20.85 \\
            w/o Warm-up & 83.1 & 55.32 \\
            \hline
            CPR & 91.1 & 79.4\\
            \bottomrule
            \end{tabular}
            }
        \label{tab5.1}
    \end{subtable}
    \hfill
    \begin{subtable}[t]{0.5\textwidth}
        \centering
        \caption{Ablation study on multi-criteria}
        {\footnotesize
        \begin{tabular}{c|c c}
            \toprule
            Ablation Study & Office(OSDA) & VisDA(OSDA)\\
            \hline
            w/o consistency condition & 84.1 & 31.03 \\
            w/o threshold condition & 88.5 & 69.1 \\
            \hline
            CPR & 91.1 & 79.4\\
            \bottomrule
            \end{tabular}
            }
        \label{tab5.2}
    \end{subtable}
    \label{tab5}
\end{table}

\subsection{Ablation study}
\textbf{The importance of $\mathcal{L}_{split}$ and warm-up.} We conduct an ablation study on split loss ($\mathcal{L}_{split}$.) and warm-up stage on Office and VisDA for OSDA (See \tabref{tab5.1}).  
If the model is firstly trained without the split loss, the performance of both experiments plummeted as shown in \tabref{tab5.1}. 
Furthermore, we can also show qualitative comparisons between with and without $\mathcal{L}_{split}$ using t-SNE (See \figref{fig4.1} and \figref{fig4.2}). As shown in \figref{fig4.2}, the model without $L_{split}$ classifies many known features as unknown class since the latent space is not well separated, which verifies that $L_{split}$ contributes greatly to  dividing feature space into known and unknown space. We also observe that some known classes are wrongly classified as unknown classes, which means reciprocal points are not well separately formed from the region of known features. Next, we can also observe that warm-up stage is a critical component of our CPR as shown in \tabref{tab5.1}. Warm-up stage seems to guarantee stable learning of two classifiers and provide an important cornerstone of multi-criteria for following phase.  

\textbf{Effectiveness of collaborative probability.} We propose reciprocal points as anchors for unknown feature space and use collaborative probability $p_c$ to classify known and unknown classes. Hence, as shown in \figref{fig3.4}, we design an anomaly score and plot the histogram of anomaly score on VisDA OSDA setting to show the effectiveness of collaborative probability. The anomaly score is  $-\log(\mathop{\max}_{K\leq j} p_c^j)$, where $p_c^j$ for $K\leq j$ is the probability of belonging to the $j$-th reciprocal point. It is a valid score due to the purpose of reciprocal points. The histogram indicates that the learned reciprocal points work well to separate unknown feature space from known one and proves the effectiveness of collaborative probability for detecting unknown feature.
\begin{figure}[t!]
    \subfloat[Analysis of $\lambda$]{{\includegraphics[width=0.24\textwidth]{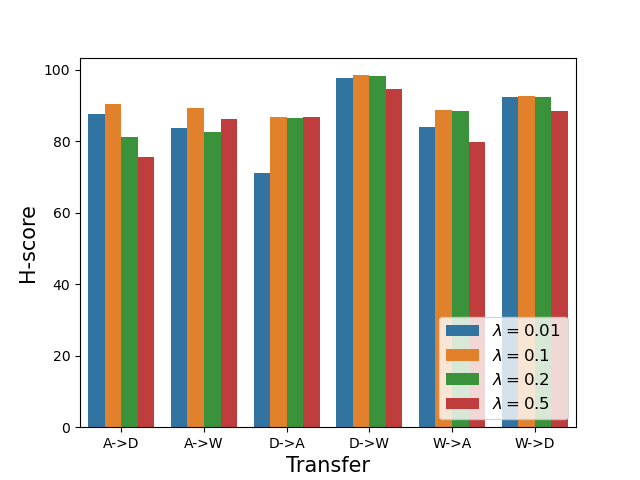}}\label{lambda}}
    \subfloat[Analysis of $i_{w}$]{{\includegraphics[width=0.24\textwidth]{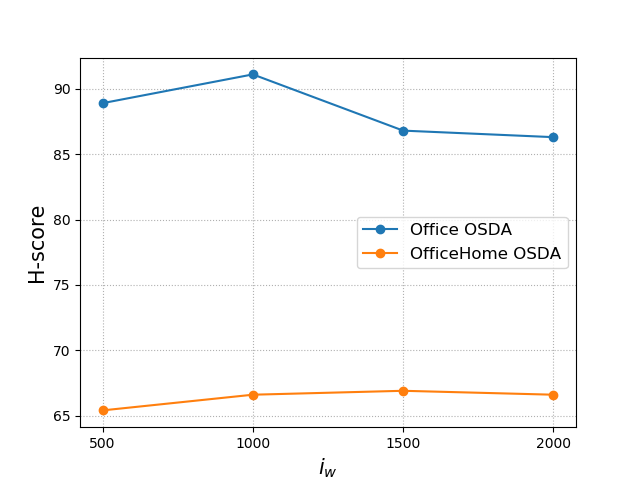}}\label{warmup}}
    \caption{(a): Analysis of $\lambda$ in the Office OSDA setting. (b): Analysis of $i_{w}$ in the Office OSDA and OfficeHome OSDA settings.}
    \label{ablation}
\end{figure}

\textbf{Analysis on multi-criteria.} Since we suggest that multi-criteria is effective strategy, we try to prove it both with quantitative and qualitative results on aforementioned two benchmark settings. First for the quantitative results as shown in \tabref{tab5.2}, if any of the condition is missing in criteria, we can easily see that the performance plunges significantly. Similarly, visual comparison between full CPR (\figref{fig4.1}) and missing conditions (\figref{fig4.3} and \figref{fig4.4}) obviously shows that only complete form of multi-criteria can result in separate space between known and unknown target samples. 

\textbf{Comparison with different $\lambda$.} We conduct experiments on Office under OSDA setting. In addition to the original model trained with $\lambda=0.1$, we also train 3 models trained with different $\lambda$ values  on the Office OSDA setting in the \figref{lambda}. The result shows the original model ($\lambda=0.1$) achieves better results for the all of scenarios and our method is robust to different choices of $\lambda$ as there is not much change in performance. 

\textbf{Analysis of $i_{w}$.} $i_{w}$ is fixed to 1000 where the thresholds are saturated. Actually, as long as thresholds are saturated, it is no matter which $i_{w}$ is used for training CPR. To show the sensitivity of CPR to the $i_{w}$, we conduct experiments on Office under OSDA and OfficeHome under OSDA for $i_w\in \{500, 1000, 1500, 2000\}$ and present average H-score for the both. As shown in \figref{warmup}, CPR is also robust to the choice of $i_{w}$. It might be obvious because $i_{w}$ is introduced to ensure the model to have enough time to get sufficiently high thresholds and reliable reciprocal points. Thus, if the model has warmed-up enough, the performance will be similar no matter when it starts adaptation phase.

\section{Conclusion}
We introduced dual Classifiers for Prototypes and Reciprocal points (CPR), a novel architecture for universal domain adaptation. This framework is motivated by the limitation of previous works that unknown samples are not properly separated from known samples without considering the underlying difference between them. We proposed a new paradigm that adopts an additional classifier for reciprocals to push them from the corresponding prototypes. To this end, our model is designed to be trained in a curriculum scheme from warm-up to adaptation stage. In warm-up stage, given the source known samples and whole target samples, we initially adapt the model with domain-specific loss. Subsequently, we utilize multi-criteria to detect confident known and unknown target samples and enhance the domain adaptation with entropy minimization on selected samples in following adaptation stage. We evaluate our model, CPR, on three and achieve comparable or new-state-of-the art results and is robustly powerful in several benchmarks with various UniDA settings. 

\section{Acknowledgement}
Institute of Information and communications Technology Planning and Evaluation (IITP) grant funded by the Korea government(MSIT) (No.2021-0-02068, Artificial Intelligence Innovation Hub).

{\small
\bibliographystyle{ieee_fullname}
\bibliography{egbib}
}

\end{document}